\documentclass{article} 
\usepackage[final]{colm2026_conference}

\usepackage{microtype}
\usepackage{url}
\usepackage{amsmath}
\usepackage{amssymb}
\usepackage{graphicx}
\usepackage{natbib}
\usepackage{booktabs}
\usepackage{enumitem}
\usepackage{xcolor}
\usepackage{hyperref}
\usepackage{tcolorbox}
\usepackage{multirow}
\usepackage{pifont}

\usepackage{lineno}

\definecolor{darkblue}{rgb}{0, 0, 0.5}
\hypersetup{colorlinks=true, citecolor=darkblue, linkcolor=darkblue, urlcolor=darkblue}

\newcommand{\A}{\mathcal{A}}

\title{Decision-Centric Design for LLM Systems}
\author{%
  {\large\fontfamily{lmr}\selectfont Wei Sun}\\[3pt]
  {\small\fontfamily{lmr}\selectfont IBM Research}\\[2pt]
  {\small\texttt{sunw@us.ibm.com}}%
}
\date{}

\begin{document}

\renewcommand{\headrulewidth}{0pt}

\makeatletter
\def\@maketitle{\vbox{\hsize\textwidth
{\LARGE\mdseries\fontfamily{lmr}\selectfont \centering \@title\par}
\vskip 0.2in
\begin{center}\@author\end{center}
\vskip 0.3in minus 0.1in}}
\makeatother

\maketitle
\lhead{}

\begin{abstract}
LLM systems must make control decisions in addition to generating outputs:
whether to answer, clarify, retrieve, call tools, repair, or escalate. In
many current architectures, these decisions remain implicit within generation,
entangling assessment and action in a single model call and making failures
hard to inspect, constrain, or repair. We propose a decision-centric framework
that separates decision-relevant signals from the policy that maps them to
actions, turning control into an explicit and inspectable layer of the system. This separation supports attribution of failures to signal estimation,
decision policy, or execution, and enables modular improvement of each
component. It unifies familiar single-step settings such as routing and
adaptive inference, and extends naturally to sequential settings in which
actions alter the information available before acting. Across three controlled
experiments, the framework reduces futile actions, improves task success, and
reveals interpretable failure modes. More broadly, it offers a general
architectural principle for building more reliable, controllable, and
diagnosable LLM systems.
\end{abstract}

\section{Introduction}
\label{sec:intro}

Language models are increasingly deployed as components of larger AI systems, where they do more than generate text: they help route requests, allocate compute, retrieve information, call tools, and participate in multi-step workflows. In these settings, system behavior depends not only on generation quality, but also on control decisions about when and how to act. Yet in many current architectures, these decisions remain implicit within generation, making failures difficult to inspect and improve.

We argue that LLM systems need an explicit decision layer. Our central design principle is simple: stochastic signals, broadly defined to include observed, contextual, and learned quantities, should be separated from the policy that selects actions. Making this interface explicit does not remove uncertainty, but it makes control inspectable. It improves traceability, enables attribution of failures to signal estimation, decision policy, or execution, and provides a natural interface for modular improvement and constraint enforcement.

We formalize this idea through a decision-centric abstraction with three elements: a set of candidate actions, a decision context comprising observed, estimated, and learned quantities, and a policy that maps context to action. This abstraction captures familiar settings such as model routing and adaptive inference scaling, and extends naturally to sequential settings in which actions can change the information available for subsequent decisions. In these settings, the framework preserves a simple interface: stochastic signals may evolve over time, but action selection remains policy-driven and explicit.

Our experiments are designed to validate complementary aspects of the framework. Section~\ref{sec:calendar} isolates the core mechanism in a minimal setting, Section~\ref{sec:graph} tests its extension to richer signals and action spaces, and Section~\ref{sec:rag} studies modular signal construction and update in a more realistic retrieval-based workflow. Across these settings, making the decision layer explicit reduces futile actions, improves task success, and exposes interpretable failure modes.

Our contributions are as follows:
\begin{enumerate}
    \item We introduce an explicit decision layer for LLM systems that separates stochastic signals from deterministic action selection, turning hidden control into an inspectable system component.
    
    \item We show that this separation exposes a common interface across both canonical control problems, such as routing and adaptive inference scaling, and sequential settings in which signals evolve over interaction.
    
    \item Across three complementary experiments, we show that making this layer explicit improves both behavior and diagnosis: it reduces futile actions, improves task success, localizes failures to signal estimation, decision policy, or execution, and provides a natural mechanism for enforcing structural constraints.
\end{enumerate}

\section{Related Work}
\label{sec:related}

\paragraph{Reasoning, search, and orchestration in LLM systems.}
A large body of work improves LLM-based systems by strengthening inference-time
reasoning or workflow orchestration. Prompting and search-based methods such as
Chain-of-Thought~\citep{wei2022chain}, Self-Consistency~\citep{wang2023selfconsistency},
Tree-of-Thoughts~\citep{yao2023tot}, and reasoning-oriented RL
approaches~\citep{guo2025deepseekr1} improve what the model generates or how much
compute is used during generation. Agentic frameworks such as
ReAct~\citep{yao2022react}, Reflexion~\citep{shinn2023reflexion},
AutoGen~\citep{wu2024autogen}, and LATS~\citep{zhou2024lats}, as well as
multi-agent architectures~\citep{hong2023metagpt,dangmulti}, instead improve
task-level orchestration through tool use, planning, reflection, and modular
decomposition. Our contribution is complementary: rather than proposing a new
prompting or orchestration strategy, we introduce an explicit decision layer for
within-task control, governing when to act, acquire information, or revise.

\paragraph{Clarification, belief modeling, and information acquisition.}
Closer to our focus is work on whether systems should gather additional
information before acting. Prior work studies clarification through
prompting~\citep{deng2023proactive,zhang2023clarifywhen},
uncertainty-triggered follow-up questions~\citep{ren2023robots,grand2025shootfirst},
and learned clarification policies~\citep{sun2025proactive,wu2025collabllm}.
Related work also models uncertainty explicitly through beliefs updated across
interaction, including probabilistic beliefs about latent user
preferences~\citep{qiu2026bayesian}, question selection by expected information
gain~\citep{choudhury2025bedllm}, and online preference elicitation through
belief updates over item utilities~\citep{austin2024bayesian}. These works show
that uncertainty estimates and belief states are meaningful objects in
LLM-mediated interaction. Our contribution differs in where control lives:
rather than leaving the act-or-clarify decision implicit in prompting or model
training, we make the decision-relevant signals and the downstream control
policy explicit and separable.

\paragraph{Decision-theoretic control in LLM systems.}
Some of the closest work already treats action selection under uncertainty as an
explicit decision problem. Canonical examples include model
routing~\citep{ong2406routellm,hu2024routerbench,tsiourvas2025causal},
adaptive inference scaling~\citep{snell2025scaling,huang2025latency}, and
adaptive retrieval control~\citep{jeong2024adaptive}, where a controller selects
 actions under quality--cost tradeoffs. Value of Information
methods~\citep{dong2026voi,raiffa1961applied,howard1966information} formalize
the clarify-or-commit tradeoff by weighing expected utility gain against
information cost. Recent Bayesian work pushes this framing further:
\citet{amin2026bayesian} formulate multi-LLM orchestration as sequential
Bayesian decision-making with value-of-information calculations, while
\citet{papamarkou2026position} argue in a position paper that agentic AI
control layers should make Bayes-consistent decisions. We share the view that control should be explicit, but our claim is
architectural rather than methodological. Bayesian control is one valid
instantiation of our framework, but not the only one: the signals driving
action selection may be Bayesian posteriors, learned uncertainty estimates,
retrieval diagnostics, structural checks, or hard domain constraints. Our
contribution is to make the separation between decision context, policy, and
execution explicit as a general design principle, and to show how that
separation extends naturally to sequential within-task control.

\section{Decision-Centric Abstraction}
\label{sec:abstraction}

We introduce a minimal abstraction that makes decision-making explicit in LLM systems by separating the information available at a decision point from the policy that selects an action.

\subsection{Core Components}

A \textbf{decision point} is a tuple $(\A, c, \delta)$:

\paragraph{Actions $\A$.}
A finite set of candidate actions available to the system, such as selecting a model, allocating inference-time compute, choosing a retrieval strategy, or taking an execution-level action.

\paragraph{Decision context $c$.}
A representation of the information relevant for action selection at a given decision point. The context may include interaction history, previous model outputs, retrieved evidence, validation outcomes, learned or heuristic signals, and hard constraints such as budgets, latency caps, or turn limits. It may be structured or unstructured; we write $c$ abstractly for simplicity.

In many current LLM systems, context interpretation, signal estimation, action selection, and output generation are entangled inside a single model call. Our framework separates these roles explicitly. An LLM may still be used inside this layer to summarize history, estimate signals, generate candidate actions, or evaluate intermediate results, but the decision interface itself is exposed. This separation improves traceability by allowing failures to be attributed to context construction, decision policy, or execution.

\paragraph{Decision function $\delta$.}
A decision function that maps context $c$ to an action in $\A$. In this paper, we focus on explicit deterministic decision functions: given the same exposed context and constraints, $\delta$ selects the same action, even if parts of that context are noisy, learned, or heuristic. The function may be rule-based, learned, or optimization-based; the key property is that action selection is explicit and repeatable.

One common instantiation is constrained utility maximization:
\begin{equation}
\delta(c)
\;=\;
a^*
\;=\;
\arg\max_{a \in \mathcal{F}(c)}
\; U(a, c),
\label{eq:decision}
\end{equation}
where $U(a,c)$ is a utility function and $\mathcal{F}(c)\subseteq \A$ denotes the feasible actions under the current context.

A common specification is a linear reward--cost tradeoff:
\begin{equation}
U(a, c)
\;=\;
R(a, c)
\;-\;
\sum_k \lambda_k C_k(a, c),
\label{eq:linear}
\end{equation}
where \(R\) denotes reward, \(C_k\) the \(k\)-th cost term, and \(\lambda_k\) the corresponding tradeoff weight. This is only one possible objective formulation; alternatives include regret-based \citep{tsiourvas2025causal}, constraint-based \citep{woisetschlaeger2025messplus}, and ranking-based \citep{somerstep2025carrot} objectives.

Making the decision interface explicit does not eliminate uncertainty: the context may be noisy, and the chosen action may still be executed by a stochastic generator. Its benefit is attribution. Once context, policy, and execution are separated, failures can be localized rather than remaining entangled inside a single model call.

\subsection{Canonical Instances: Routing and Inference Scaling}

This abstraction already appears in several familiar settings. Two canonical examples are model routing and adaptive inference scaling. Table~\ref{tab:single_turn_mapping} highlights the shared structure.

\begin{table}[t]
\centering
\small
\begin{tabular}{@{}p{2.8cm}p{4.8cm}p{5.0cm}@{}}
\toprule
\textbf{Component} & \textbf{Model Routing} & \textbf{Adaptive Inference Scaling} \\
\midrule
Actions $\mathcal{A}$
& Candidate models
& Inference strategies and compute configurations \\[4pt]

Decision context $c$
& Query features, predicted quality, API cost, latency, SLA
& Query features, difficulty signals, token budget, latency constraints \\[4pt]

Decision problem $\delta(c)$
& \multicolumn{2}{c}{$\arg\max_{a \in \mathcal{F}(c)} U(a,c)$, where $U$ trades quality against cost (Eq.~\ref{eq:decision})} \\
\bottomrule
\end{tabular}
\caption{Two canonical settings mapped to the decision-centric abstraction ($\mathcal{A}$, $c$, $\delta$). Both instantiate the same structure; they differ in what actions represent and what signals populate the context.}
\label{tab:single_turn_mapping}
\end{table}

In \textbf{model routing}, the action is which model to invoke, and the utility may reflect quality--cost tradeoffs, regret, or service-level objectives~\citep{woisetschlaeger2025messplus,tsiourvas2025causal}. In \textbf{adaptive inference scaling}, the action is which inference strategy to use and how much computation to allocate, for example through the number of samples, voting rounds, or search depth~\citep{damani2025learning,huang2025latency}. 
These settings share the same architectural pattern: an explicit action space, a decision context, and a policy over feasible actions. In both, the context is treated as given at the moment of choice. The next section turns to sequential settings, where actions can change the information available to later decisions.

\section{Sequential Decision-Making with Evolving Context}
\label{sec:sequential}

In sequential settings, systems repeatedly face control decisions about how to proceed from the current context. In the minimal case, this reduces to deciding whether to act immediately or seek more information. Existing approaches typically handle even this simple form of control implicitly.

\subsection{A Minimal Sequential Formulation}

In sequential settings, systems often face a recurring decision: act on the current context or
seek more information. Existing approaches usually handle this implicitly. Retry loops act,
validate, and try again after failure; prompt-based methods ask the model to decide within a
single generation whether to execute or clarify. In both cases, control remains entangled with
generation, making it difficult to inspect the basis for the decision or attribute failures to
signal estimation, policy, or execution.

In the decision-centric formulation, the decision context at turn $t$, denoted $c_t$,
contains the decision-relevant information available at that point, including the original
request, prior model outputs, user responses, retrieved evidence, and validation outcomes.
The framework does not prescribe a particular representation of this context. Its only
architectural requirement is that whatever drives action selection be exposed explicitly,
rather than left implicit inside a generation call.

In the minimal instantiation used here, action selection depends on a single explicit signal:
a sufficiency score $\hat{p}_{\mathrm{suff},t} \in [0,1]$, estimated from $c_t$, that reflects
whether the current context appears sufficient to act reliably. In our experiments, this
signal is produced by an LLM-based estimator operating directly on unstructured inputs. More
generally, the same role could be played by rule-based checks, structural completeness
measures, retrieval scores, or learned estimators over structured or unstructured features.

A policy then maps the exposed signal to an action. For example, a simple threshold rule may
choose \texttt{execute} when $\hat{p}_{\mathrm{suff},t}$ exceeds a preset threshold and
\texttt{clarify} otherwise. More generally, the policy may be learned, optimization-based, or
constraint-driven. The key point is that control is explicit: given the same exposed signal,
the policy selects the same action. The chosen action then produces new observations, which
update $c_t$ for the next turn.

This separation makes failures attributable. If the system acts too early, the source of
error can be traced to the sufficiency estimate, the policy, or the execution itself. It also
allows structural constraints to be enforced directly at the decision level. In
Section~\ref{sec:calendar}, for example, a failed \texttt{execute} is followed by
\texttt{clarify} rather than blind retry, preventing wasted turns that do not acquire new
information.

\subsection{A Multi-Signal Instantiation}
\label{sec:beliefs}

The minimal formulation uses a single explicit signal, but the same interface also supports settings in which multiple decision-relevant quantities are exposed separately and used jointly by the policy. In this paper, we study one such instantiation with two explicit signals. This is only one realization of the broader framework: the decision layer does not require belief-based signals, only that the quantities driving action selection be made explicit.

Concretely, we use a sufficiency signal, $\hat{p}_{\mathrm{suff},t} \in [0,1]$, which reflects whether the current context contains enough information to act reliably, and a correctness signal, $\hat{p}_{\mathrm{corr},t} \in [0,1]$, which reflects whether the current trajectory or intermediate result appears to be on the right track. Low sufficiency favors information acquisition before acting; low correctness after an attempted action favors revision or backtracking.

The value of exposing both signals is that the optimal action can depend on their \emph{joint} state, and actions can update beliefs they were not intended to target. A failed attempt may lower perceived correctness while simultaneously improving sufficiency by eliminating alternatives. This multi-signal view supports richer action spaces without changing the architectural separation between context, policy, and execution. Section~\ref{sec:graph} instantiates this setting in a graph disambiguation task, where sufficiency and correctness co-evolve across turns and jointly determine whether the system should clarify, continue, or backtrack.

\section{Experiments}
\label{sec:exp_setup}

We evaluate the decision-centric framework through three experiments, each isolating a distinct property of explicit control. Section~\ref{sec:calendar} studies the minimal act-versus-clarify decision under a single sufficiency signal. Section~\ref{sec:graph} extends this setting to multiple co-evolving signals and a richer action space. Section~\ref{sec:rag} tests modularity in retrieval control by holding the decision rule fixed and varying only the construction of the sufficiency signal.  

Unless otherwise specified, all experiments use \texttt{ibm/granite4:micro}~\citep{granite2025micro}, run locally via Ollama\footnote{\url{https://ollama.com}}. We compare a \textbf{Prompt} baseline, which leaves action selection implicit in a single LLM call, with the \textbf{Decision-Centric} approach (\textbf{DC}), which computes explicit signals and applies a deterministic policy; where relevant, we also include a \textbf{Retry} baseline that regenerates after failure without an explicit action decision.

\subsection{Calendar Scheduling: Clarify vs.\ Execute}
\label{sec:calendar}

\paragraph{Setup.}
The task is to produce a valid calendar event from a natural-language request
with four required fields: date, start time, duration, and attendees. Requests
may be incomplete or ambiguous. We vary both the number of missing fields,
$k \in \{0,1,2,3,4\}$, and the ambiguity type: \emph{absent}, where a field is
omitted, and \emph{unresolvable}, where a field is mentioned but not specified
in usable form (e.g., ``Schedule on Jack's usual slot''). 

At each turn, the system chooses between \texttt{clarify} and
\texttt{execute}. In DC, this choice is driven by an explicit binary
sufficiency signal indicating whether all required fields are available in
usable form: if sufficient, the system executes; otherwise, it asks a
clarification question. We compare Prompt-Clarify, Retry, and DC over
$N{=}10$ runs per scenario, and report success, first-action optimality,
wasted executions, and clarification turns. Additional implementation details
are provided in Appendix~\ref{app:implementation} - ~\ref{app:prompt_clarify}.

\begin{table}[t]
\centering
\small
\begin{tabular}{c cccc | cccc}
\toprule
& \multicolumn{4}{c|}{Prompt-Clarify} & \multicolumn{4}{c}{DC (ours)} \\
\cmidrule(lr){2-5}\cmidrule(lr){6-9}
$k$ & Succ. & 1st & Wasted & Clarif. & Succ. & 1st & Wasted & Clarif. \\
\midrule
0 & 100\% &  90\% & 0.00 & 0.10 & 100\% & 100\% & 0.00 & 0.00 \\
1 & 100\% &  10\% & 1.35 & 1.25 & \textbf{100\%} & \textbf{100\%} & \textbf{0.65} & \textbf{1.65} \\
2 &  75\% &  10\% & 2.20 & 2.25 & \textbf{100\%} & \textbf{100\%} & \textbf{0.00} & \textbf{1.00} \\
3 &  60\% &  45\% & 2.20 & 2.80 & \textbf{100\%} & \textbf{100\%} & \textbf{0.55} & \textbf{1.55} \\
4 &  10\% &  10\% & 2.90 & 3.00 & \textbf{100\%} & \textbf{100\%} & \textbf{0.10} & \textbf{1.40} \\
\bottomrule
\end{tabular}
\caption{Calendar results by missing-field count $k$, averaged over ambiguity types. \emph{1st}: first-action optimality; \emph{Wasted}: discarded executions; \emph{Clarif.}: clarification turns per run. }
\label{tab:results_by_missing}
\end{table}

\paragraph{Results.}
Table~\ref{tab:results_by_missing} shows the main  results on Granite.
The decision-centric system achieves 100\% success in all 8 scenarios,
while Prompt-Clarify degrades sharply as missingness increases,
dropping to 75\%, 60\%, and 10\% at $k{=}2,3,4$, respectively. DC also
incurs fewer wasted executions and uses fewer turns, showing that the
gain comes from better control decisions rather than more favorable
generation.
The Retry baseline succeeds only at $k{=}0$ and fails at all $k{\geq}1$
by exhausting the turn budget on repeated executions without ever clarifying;
it is omitted from Table~\ref{tab:results_by_missing} for space.

The full per-scenario breakdown
is given in Table~\ref{tab:appendix_full}. The gap is largest in unresolvable scenarios, where missing fields are
referenced but not inferable. There,
Prompt-Clarify often treats the reference as usable information
and executes prematurely, whereas DC continues clarifying until the
missing fields are explicitly resolved.

We replicate the same experiment on LLaMA~3~(8B)~\citep{llama3_2024}; results are shown in Table~\ref{tab:cross_model_main}. The main pattern transfers across model families: 
DC continues to outperform both Retry and Prompt-Clarify at every
missing-field count $k$, but the error pattern becomes non-monotonic,
with lower performance at $k{=}1$ than at $k{=}2$.

Because the architecture separates estimation, policy, and execution,
this irregularity can be localized rather than treated as an opaque
model effect.
In the main $k{=}1$ failures, the sufficiency estimator correctly
identifies that only \texttt{duration min} is missing and the policy
correctly selects \texttt{clarify}, but the question generator asks
about already known fields (e.g., \emph{``What time should the meeting
start?''}).
This prevents the system from acquiring the missing information within
the turn budget.

Constraining the question generator to ask only about fields identified
as missing resolves these failures, restoring correct behavior without
modifying the estimator, the controller, or the executor.
This illustrates a core benefit of the framework: failures are
attributable to specific components, and fixes can be applied locally
without changing the overall decision logic. Full traces
(Figure~\ref{fig:llama_trace}) and additional failure analysis are in
Appendix~\ref{app:cross_model}.

\begin{table}[h]
\centering
\small
\begin{tabular}{c rrrr}
\toprule
$k$ & Retry & Prompt-Clarify & DC (original) & DC (constrained)$^\dagger$ \\
\midrule
0 & 100\% &  70\% & 100\% & 100\% \\
1 &   5\% &  50\% &  75\% & 100\% \\
2 &   0\% &  75\% & 100\% & 100\% \\
3 &   0\% &  45\% &  90\% &  95\% \\
4 &   0\% &  30\% & 100\% & 100\% \\
\bottomrule
\end{tabular}
\caption{LLaMA success rates by missing-field count $k$, averaged over ambiguity types. DC~(original) uses the default question-generation module; $^\dagger$DC~(constrained) restricts question generation to the listed missing fields.}
\label{tab:cross_model_main}
\end{table}

\subsection{Graph Disambiguation: Joint Beliefs and Correlated Updates}
\label{sec:graph}

\paragraph{Setup.}
We instantiate the joint-belief setting of
Section~\ref{sec:beliefs}, in which both $\hat{p}_\text{suff}$ and
$\hat{p}_\text{corr}$ are active and the optimal action depends on their joint
state. The domain is a synthetic knowledge graph with 200 people, each
described by five categorical attributes: department, role, location, project,
and level. Given a \emph{partial description} of a target person, the system
must identify the correct node by interleaving three actions:
\textbf{clarify} (ask for a missing attribute), \textbf{execute} (visit a
candidate and observe its full profile), and \textbf{backtrack} (reject the
current candidate and try a different one from the remaining pool). For example, the query ``Find the
person in Engineering, project Alpha, whose role is Manager'' may still match
multiple candidates, requiring the system to resolve the ambiguity within a
fixed turn budget.

We evaluate five scenarios that isolate and then couple the two signals.
S1--S3 are controls that separate sufficiency and correctness, while S4--S5
introduce joint-belief settings in which the optimal action depends on their
interaction, including cases where belief updates arise indirectly through
traversal. Details of the graph generation process and scenario design are in Appendix~\ref{app:graph_task}.

In addition to Retry, Prompt, and DC, we include a \textbf{Prompt (w/ policy)}
condition, which states the decision logic in the system prompt rather than
implementing it as an explicit decision function. This tests whether the same
policy can be executed reliably through prompting alone.
All methods receive the same turn-level state: the query, known and unknown
attributes, the number of remaining and untried candidates, the last visited
profile, and the remaining budget. The difference is purely architectural. DC
computes explicit beliefs and applies a deterministic policy, whereas Prompt,
Prompt (w/ policy), and Retry must infer the next action from the same state.

For DC, the sufficiency signal
$\hat{p}_\text{suff}=1/|\text{candidates}|$ measures query ambiguity and
increases as the candidate set shrinks, either through clarification or through
elimination after failed traversal. The correctness signal
$\hat{p}_\text{corr}$ is estimated after each visit from local structural
consistency: for each hidden attribute, we compute the fraction of peer
candidates that share the visited node's value. Rare values reduce plausibility,
whereas common values increase it. Full implementation details are given in
Appendix~\ref{app:graph_dc} and Appendix~\ref{app:graph_prompts}.

\begin{table}[t]
\centering
\small
\begin{tabular}{@{}l rrrr@{}}
\toprule
Scenario & Retry & Prompt & \shortstack{Prompt\\(w/ policy)} & \textbf{DC} \\
\midrule
S1: Clean (baseline)                       & 100\% & 100\% & 100\% & \textbf{100\%} \\
S2: Ambiguous ($\hat{p}_\text{suff}$ only) &  45\% &  85\% &  95\% & \textbf{100\%} \\
S3: Unreliable ($\hat{p}_\text{corr}$ only)& 100\% & 100\% & 100\% & \textbf{100\%} \\
S4: Orthogonal (joint, $T{=}6$)            &  65\% &  95\% & 100\% & \textbf{100\%} \\
S5: Correlated (coupled update)            &  60\% &  35\% &  35\% & \textbf{100\%} \\
\bottomrule
\end{tabular}
\caption{Graph disambiguation success rates.}
\label{tab:graph_results}
\end{table}

\paragraph{Results.}
Table~\ref{tab:graph_results} summarizes the main results. DC achieves
100\% success in all scenarios, while the baselines degrade as the task
requires more complex control. A full breakdown, including wasted
clarifications and backtracks, is given in
Table~\ref{tab:graph_full}.

\paragraph{Control scenarios (S1--S3).}
S1 is trivial: all methods succeed, confirming that the pipeline functions
end-to-end. S2 isolates sufficiency. With 13 candidates and only two known
attributes, DC reaches 100\% by clarifying until the pool is small enough to
act on. Retry drops to 45\% because it executes blindly, while Prompt-clarify
reaches 85\%, occasionally acting too early without explicit sufficiency
tracking. Prompt (w/ policy) improves to 95\%, indicating that policy prose can
help when the correct action follows directly from the observable candidate
count. S3 isolates correctness. Here all methods succeed, since with only one
untried candidate remaining, even Retry eventually recovers by exhaustion.

\paragraph{Joint belief: action dependence (S4).}
S4 is the first joint-belief setting. After a forced decoy traversal fails, the
correct next action depends on both signals: $\hat{p}_\text{corr}$ indicates
rejection of the visited node, but $\hat{p}_\text{suff}$ determines whether the
system should backtrack or clarify. Because many candidates remain, DC
clarifies and achieves 100\%. Prompt-clarify reaches 95\%, failing when it
backtracks instead of clarifying. Prompt (w/ policy) also reaches 100\%,
showing that prompt-level policy description is sufficient when the
post-failure decision can still be read directly from the candidate state.

\paragraph{Correlated belief update (S5).}
S5 is the strongest test. After an initial clarification reduces the pool from
12 to 5 candidates, DC executes with one hidden attribute remaining. The forced
decoy then fails, but this failure also eliminates other candidates sharing the
decoy's minority hidden value, shrinking the pool from 5 to 3 without any
additional clarification. DC updates its belief state in place, backtracks into
the reduced pool, and reaches 100\%.

The prompt-based methods fail for a different reason. Prompt-clarify reaches
35\%, and Prompt (w/ policy) remains at 35\%, even below the Retry baseline at
60\%. Unlike S2 and S4, the bottleneck is not deciding how to act on an
observable candidate count. Rather, the effective candidate set has already
changed during traversal, and that state update is not available to the LLM at
decision time. DC succeeds because it maintains and updates the belief state
explicitly; embedding the same policy in prompt text does not recover that
missing state.

S4 and S5 thus extend the calendar setting in two ways: the correct action can
depend on multiple explicit beliefs, and actions can update those beliefs across
turns through their effects on the candidate state.

\subsection{Modular Signals for Retrieval Control with Diagnosable Feedback}
\label{sec:rag}

While the first two experiments use controlled settings to isolate decision behavior, we next evaluate the same design in a more realistic retrieval setting, where signal quality is imperfect and decisions must be made under noisy evidence.

\paragraph{Setup.}
We study a \emph{retrieval sufficiency} task, in which the system must decide
whether the currently retrieved passage set is sufficient to answer a question.
At each turn, it observes the question, the accumulated BM25 passages, and the
remaining retrieval budget, and chooses either \texttt{stop} or
\texttt{expand\_k}, with $k=3 \rightarrow 6 \rightarrow 9$. Success is defined
by whether the gold answer string is present in the final passage set at stop
time. We also report the average number of retrieval rounds per episode, which
captures decision efficiency: strong methods stop early on easy questions while
continuing to expand on medium ones. The retrieval budget is two rounds; once
exhausted, the controller must \texttt{stop}.

We evaluate 150 Natural Questions examples~\citep{kwiatkowski2019natural} over
a 2000-passage BM25 corpus, split evenly into three buckets ($N=50$ each) by
when the gold passage becomes retrievable: \textbf{easy} (present at round~0),
\textbf{medium} (absent at round~0 but retrieved within budget), and
\textbf{hard} (not retrieved within budget). Retrieval states are precomputed
once and shared across all methods. Corpus construction and bucket definitions
are given in Appendix~\ref{rag:setup}.

We compare four methods: \emph{Prompt}, \emph{DC-LLM}, \emph{DC-Dense}, and
\emph{DC-Composite}. \emph{Prompt} maps the current state directly to
\texttt{stop} or \texttt{expand\_k} in a single LLM call, leaving sufficiency
assessment and action selection implicit. The three decision-centric variants
instead externalize sufficiency as an explicit scalar signal
$\hat{p} \in [0,1]$, interpreted as the probability that the current passage set
is sufficient. \emph{DC-LLM} derives $\hat{p}$ from an LLM-based answerability
judge; \emph{DC-Dense} derives it from question--passage embedding similarity
using \texttt{all-MiniLM-L6-v2}; and \emph{DC-Composite} combines the two.

All methods receive the same state and act over the same action set. The three
DC variants also share the same fixed threshold controller and differ only in
how $\hat{p}$ is constructed. This isolates signal quality from control logic:
performance differences among DC variants reflect the quality of the sufficiency
signal rather than differences in action generation. Signal definitions are
given in Appendix~\ref{rag:signals}.

\paragraph{Results. }

\begin{table}[t]
\centering
\small
\begin{tabular}{l rr rr rr rr}
\toprule
& \multicolumn{2}{c}{Prompt} & \multicolumn{2}{c}{DC-LLM} & \multicolumn{2}{c}{DC-Dense} & \multicolumn{2}{c}{DC-Composite} \\
\cmidrule(lr){2-3}\cmidrule(lr){4-5}\cmidrule(lr){6-7}\cmidrule(lr){8-9}
Bucket & Succ. & RR & Succ. & RR & Succ. & RR & Succ. & RR \\
\midrule
Easy   & 100\% & 0.10 & 100\% & 0.62 & 100\% & 1.12 & 100\% & 0.84 \\
Medium &  14\% & 0.08 &  88\% & 1.34 &  90\% & 1.66 &  94\% & 1.62 \\
Hard   &  14\% & 0.14 &  18\% & 1.66 &  18\% & 1.84 &  18\% & 1.84 \\
\bottomrule
\end{tabular}
\caption{Retrieval control results.
  Succ.\ = success rate; RR = avg retrieval rounds.}
\label{tab:rag_results}
\end{table}

\emph{Prompt} and \emph{DC-LLM} use the same LLM, retrieved passages, and action
set; the only difference is whether sufficiency is externalized as an explicit
scalar signal. On medium questions, where the stop-versus-expand decision is
consequential, \emph{Prompt} reaches 12\% success, while \emph{DC-LLM} reaches
86\%. In 64\% of \emph{Prompt}'s medium failures, the model explicitly states
that the answer is absent from the current passages yet still chooses
\texttt{stop}, illustrating the core failure of implicit control: the
assessment is correct, but the action is wrong.

The main contribution of this experiment, however, is modularity. The three DC
variants share the same controller and differ only in how the sufficiency signal
$\hat{p}$ is constructed. \emph{DC-Dense} behaves like a relevance signal and
tends to over-expand; \emph{DC-LLM} behaves more like an answerability signal
and is more selective; \emph{DC-Composite} combines the two and achieves the
best medium-bucket performance at 94\%. Because the controller is fixed, these
differences isolate the effect of signal construction.

Logging $\hat{p}$ as a scalar trace also makes failures directly diagnosable and
supports modular iteration: new signals can be introduced, compared, and
evaluated under the same controller without changing the decision policy.

\paragraph{Takeaway.}
Taken together, the three experiments show that the value of an explicit
decision layer is not tied to a single task or signal type. In calendar
scheduling, it reduces wasted actions by enforcing simple sufficiency-based
control. In graph search, it supports multi-belief, cross-turn decisions that
implicit prompting does not reliably recover. In retrieval, it makes sufficiency
a modular and diagnosable signal, enabling attribution and improvement without
changing the controller. The common benefit is architectural: once
decision-relevant state is externalized, control becomes more reliable,
interpretable, and easier to improve.

\section{Conclusion}

We introduced a decision-centric framework for LLM systems that separates
explicit decision-relevant signals from the policy that maps them to actions.
This turns control from an implicit byproduct of generation into an explicit,
inspectable, and enforceable part of the system.

Across three experiments, this separation improved different aspects of control.
In calendar scheduling, it reduced wasted execution by making
clarify-versus-execute decisions explicit. In graph disambiguation, it enabled
richer sequential control with multiple signals, including backtracking when the
current trajectory became unreliable. In retrieval, it made stopping decisions
modular, allowing different sufficiency signals to be substituted or combined
without changing the controller.

As models become more capable, some of the gaps we observe may narrow, and
prompt-based systems may improve with stronger models or heavier prompt
engineering. But the underlying limitation remains: prompt-based control is
implicit. Assessment and action are fused inside a single model call, making the
basis for behavior hard to inspect, enforce, diagnose, or repair, especially in
sequential settings where errors compound over time.

This is where the decision-centric framework matters. By separating decision
context, policy, and execution, it provides explicit control points for
attribution, modular redesign, and constraint enforcement. These properties are
useful not only when models are weak, but also as LLM systems become more
capable and more complex.

Natural extensions include hierarchical decision layers, where a high-level
policy selects a broad action type and a lower-level policy determines how to
realize it. The same abstraction may also apply within larger multi-step
workflows, including agent-controlled ones, where each task exposes its own
local signals and policy before execution proceeds. More broadly, we view
explicit separation between decision context, policy, and execution as a useful
principle for building more reliable, controllable, and diagnosable LLM
systems.

\bibliographystyle{plainnat}
\bibliography{references}

\appendix 


\section{Calendar Scheduling: Clarify vs. Execute}
\label{app:calendar}
\subsection{Experimental Setup} \label{app:implementation}

\paragraph{Task.}
The system must produce a valid structured output, i.e., a calendar event
with four required fields (date, start\_time, duration\_min,
attendees), from a natural-language request that may be incomplete
or ambiguous. Success requires all four fields to be correct;
fabricating a missing field does not count as success.

\paragraph{Scenario Design}
\label{app:scenarios}
Scenarios are generated programmatically from a single base fact set
($\texttt{date}{=}$2026-02-17, $\texttt{start\_time}{=}$11:30, $\texttt{duration\_min}{=}$30, $\texttt{attendees}{=}$[Jack])
by varying two independent factors:

\begin{itemize}[leftmargin=*,itemsep=2pt]
  \item \textbf{Missing field count} $k \in \{0,1,2,3,4\}$: the number of
    fields withheld from the initial query, chosen in a fixed omission order
    (\texttt{duration\_min} $\to$ \texttt{date} $\to$ \texttt{start\_time} $\to$ \texttt{attendees})
    so that each $k$ produces a canonically distinct scenario.
  \item \textbf{Ambiguity type}: \emph{absent} (field simply omitted) or
    \emph{unresolvable} (field referenced but not inferable, e.g.\
    ``on Jack's usual slot'' for \texttt{date}).
    For the unresolvable type, the first observed field in a fixed priority
    order is replaced with a vague reference.
\end{itemize}

The $k{=}0$ case has all fields present; ambiguity type does not apply.
The $k{=}4$ unresolvable case is identical to $k{=}4$ absent (no fields
remain to make unresolvable), so it is excluded.
This yields $1 + 3{\times}2 + 1 = 8$ unique scenarios.
Table~\ref{tab:scenarios} shows all eight scenarios with their queries.

\begin{table}[h]
\centering
\small
\begin{tabular}{c l p{7.5cm}}
\toprule
$k$ & Ambiguity & Initial query \\
\midrule
0 & --- & Schedule a meeting with Jack on 2026-02-17 at 11:30 for 30 minutes. \\
\midrule
1 & Absent      & Schedule a meeting with Jack on 2026-02-17 at 11:30. \\
1 & Unresolvable & Schedule a meeting with Jack on Jack's usual slot at 11:30. \\
\midrule
2 & Absent      & Schedule a meeting with Jack at 11:30. \\
2 & Unresolvable & Schedule a meeting with Jack at Jack's usual time. \\
\midrule
3 & Absent      & Schedule a meeting with Jack. \\
3 & Unresolvable & Schedule a meeting with the usual team. \\
\midrule
4 & --- & Schedule a meeting. \\
\bottomrule
\end{tabular}
\caption{All 8 generated scenarios. Fields withheld at each $k$:
  $k{=}1$: \texttt{duration\_min};
  $k{=}2$: \texttt{duration\_min}, \texttt{date};
  $k{=}3$: \texttt{duration\_min}, \texttt{date}, \texttt{start\_time};
  $k{=}4$: all fields.
  Private facts (ground truth) are identical across all scenarios.}
\label{tab:scenarios}
\end{table}

We set the turn budget to $T{=}6$, the smallest budget under which the
hardest case can still be solved without mistakes: in the worst case,
all four missing fields must be clarified before a final execution,
yielding a six-turn horizon. This makes the budget tight enough to penalize
premature or wasted executions while still allowing an optimal policy to
succeed. Each scenario is repeated for $N{=}10$ independent runs.

\paragraph{Metrics.}
\textbf{Success rate} measures whether the final output is correct.
\textbf{First-action optimality} measures whether the system's first action matches the
optimal (execute at $k{=}0$; clarify at $k{\geq}1$); it isolates the act-vs.-clarify
decision quality before any recovery can occur.
\textbf{Wasted executions} counts generation calls whose outputs were discarded because
the system executed before it had sufficient information---a direct measure of decision
quality independent of output quality.
\textbf{Clarifications} counts how many turns were spent acquiring information:
DC bundles all missing fields into one question, while Prompt may ask redundantly
or execute prematurely.


\paragraph{Shared Execution Prompt}
\label{app:exec_prompt}

All three approaches use the same execution prompt to generate the final calendar event JSON
(temperature~$0.2$). This isolates the decision mechanism as the sole variable across approaches:

\begin{quote}\small\ttfamily
You create calendar event JSON. Today is \{CURRENT\_DATE\}.\\[4pt]
Required fields: date (YYYY-MM-DD), start\_time (HH:MM), duration\_min (integer), attendees (list of strings).\\
- Scan the ENTIRE conversation for each field and use the most recently confirmed value.\\
- Include ONLY values the user explicitly provided --- set missing fields to null (do NOT invent values).\\
- For attendees: if any person's name appears anywhere in the conversation, include them.\\
- attendees MUST be a JSON array of plain strings, e.g.\ ["Jack"] not [\{"name":"Jack"\}].\\
Return ONLY JSON with exactly those four keys.
\end{quote}

\paragraph{User Simulator}
\label{app:user_sim}

The user simulator role-plays as a user who holds the private ground-truth facts.
When asked a clarification question, it looks up the answer in the private facts
and generates a natural-language reply (temperature~$0.3$), answering only what
was asked without volunteering extra information.
For unresolvable scenarios (e.g., date referenced as ``Jack's usual slot''),
the simulator still knows the true date and will provide it if asked directly ---
the unresolvability lies in the \emph{initial query phrasing}, not in the simulator's knowledge.
This means the system can always recover from an unresolvable reference by asking.

\subsection{Decision-Centric Baseline Implementation}\label{app:dc_implementation}

\paragraph{Policy.}
The decision rule selects $a_t \in \{\texttt{execute}, \texttt{clarify}\}$
using a deterministic three-branch policy with no tunable parameters:

\begin{tcolorbox}[colback=gray!6,colframe=gray!40,boxrule=0.4pt,
  title={\small\textbf{Decision Rule}},fonttitle=\small,
  left=4pt,right=4pt,top=3pt,bottom=3pt]
\small\ttfamily
\textbf{Input:}\ $\hat{p}_{\mathrm{suff},t}$,\ last\_action,\ last\_val\\[2pt]
1.\ \textbf{if}\ last\_action\ ==\ execute\ \textbf{and}\ last\_val.valid\ ==\ False:\\
\phantom{1.\ }$\rightarrow$\ \textbf{CLARIFY}\hfill{\normalfont\itshape(no-blind-retry)}\\[2pt]
2.\ \textbf{elif}\ $\hat{p}_{\mathrm{suff},t}$\ ==\ 1.0:\\
\phantom{1.\ }$\rightarrow$\ \textbf{EXECUTE}\hfill{\normalfont\itshape(all fields confirmed)}\\[2pt]
3.\ \textbf{else}:\\
\phantom{1.\ }$\rightarrow$\ \textbf{CLARIFY}\ {\normalfont(ask about all unconfirmed fields at once)}
\end{tcolorbox}

The policy is explicit, deterministic, and fully attributable:
given the same $\hat{p}_{\mathrm{suff},t}$ and action history,
it always selects the same action.

Branch~1 is a hard structural constraint: after a failed execution the system
\emph{must} clarify, regardless of the current signal value.
This kind of constraint is straightforward to enforce in an explicit decision
layer --- it is a single guard before the policy branches --- but is difficult
to enforce reliably in a prompt-based system, where the model may choose to
retry directly without any mechanism to prevent it.
Neither baseline imposes this constraint.

\paragraph{Sufficiency estimator}
The estimator makes a single LLM call on the full conversation history
at each turn (temperature~$0.1$) and returns a boolean presence indicator
per field rather than a scalar probability:

\begin{quote}\small\ttfamily
Check which scheduling fields were explicitly provided by the user in this conversation.\\
Return ONLY JSON: \{"date": true/false, "start\_time": true/false, "duration\_min": true/false, "attendees": true/false\}\\[4pt]
- date: true if an explicit date or computable relative date appears (e.g.\ "2026-02-17", "next Tuesday"). false if missing or unresolvable ("Jack's usual slot").\\
- start\_time: true if a specific time appears ("11:30", "11am", "14:30"). false if missing.\\
- duration\_min: true if a numeric duration appears ("30 minutes", "1 hour", "for 45 min"). false if vague or missing.\\
- attendees: true if any person is named. false if missing.
\end{quote}

The sufficiency score is then computed by code as $\hat{p}_t = |\text{confirmed fields}| / 4$.
A \emph{confirmed-field lock} ensures that once a field is marked \texttt{true},
it remains confirmed in all subsequent turns regardless of later extractor outputs,
preventing regression as the conversation grows.
The full history is passed on every turn so the model can resolve
short or ambiguous answers (e.g., ``45'' is correctly identified as
\texttt{duration\_min} given the preceding ``How long?'' question).

\paragraph{Clarification question generation}

When the policy selects \texttt{clarify}, a single LLM call generates one natural
question covering all unconfirmed fields (temperature~$0.3$):

\begin{quote}\small\ttfamily
You are a scheduling assistant. Today is \{CURRENT\_DATE\}.\\
You are collecting information to book a calendar event.\\
Generate ONE natural question that asks the user for all the specified missing fields.\\
Return ONLY JSON: \{"question": "..."\}
\end{quote}

The missing fields are passed as input (e.g., \texttt{["date", "duration\_min"]}),
and the model generates a single fluent question covering all of them.

\subsection{Prompt-Clarify Baseline Implementation}\label{app:prompt_clarify}
The following minimal system prompt is used by the \emph{prompt-based clarify baseline only}
(temperature~$0.5$). 

\begin{quote}\small\ttfamily
You are a scheduling assistant. Today is \{CURRENT\_DATE\}.\\
You are helping book a calendar event. You need: date, start time, duration, and attendees.\\
At each turn, choose one action:\\
\quad \{"action": "clarify", "question": "..."\} --- ask the user for anything missing or unclear\\
\quad \{"action": "execute"\} --- when you have everything needed to book the event
\end{quote}

%
\subsection{Full Results by Ambiguity Type (Granite~4)}
\label{app:full_scenarios}

Table~\ref{tab:appendix_full} gives the per-ambiguity-type breakdown underlying Table~\ref{tab:results_by_missing}, adding average turns as an efficiency metric (Granite~4 micro, $T{=}6$, $N{=}10$).

\begin{table}[h]
\centering
\small
\begin{tabular}{c l l rrr}
\toprule
$k$ & Ambiguity & Approach & Success & Wasted & Avg turns \\
\midrule
\multirow{3}{*}{0}
& \multirow{3}{*}{---}
& Retry baseline       & 100\% & 0.00 & 1.00 \\
&& Prompt-based clarify & 100\% & 0.00 & 1.10 \\
&& Decision-centric     & 100\% & 0.00 & 1.00 \\
\midrule
\multirow{6}{*}{1}
& \multirow{3}{*}{Absent}
& Retry baseline       &   0\% & 6.00 & 6.00 \\
&& Prompt-based clarify & 100\% & 0.90 & 2.90 \\
&& Decision-centric     & 100\% & 0.10 & 2.20 \\
\cmidrule{2-6}
& \multirow{3}{*}{Unresolvable}
& Retry baseline       &   0\% & 6.00 & 6.00 \\
&& Prompt-based clarify & 100\% & 1.80 & 4.80 \\
&& Decision-centric     & 100\% & 1.20 & 4.40 \\
\midrule
\multirow{6}{*}{2}
& \multirow{3}{*}{Absent}
& Retry baseline       &   0\% & 6.00 & 6.00 \\
&& Prompt-based clarify & 100\% & 1.90 & 4.90 \\
&& \textbf{Decision-centric} & \textbf{100\%} & \textbf{0.00} & \textbf{2.00} \\
\cmidrule{2-6}
& \multirow{3}{*}{Unresolvable}
& Retry baseline       &   0\% & 6.00 & 6.00 \\
&& Prompt-based clarify &  50\% & 2.50 & 6.00 \\
&& \textbf{Decision-centric} & \textbf{100\%} & \textbf{0.00} & \textbf{2.00} \\
\midrule
\multirow{6}{*}{3}
& \multirow{3}{*}{Absent}
& Retry baseline       &   0\% & 6.00 & 6.00 \\
&& Prompt-based clarify &  60\% & 2.20 & 5.60 \\
&& \textbf{Decision-centric} & \textbf{100\%} & \textbf{0.00} & \textbf{2.00} \\
\cmidrule{2-6}
& \multirow{3}{*}{Unresolvable}
& Retry baseline       &   0\% & 6.00 & 6.00 \\
&& Prompt-based clarify &  60\% & 2.20 & 5.60 \\
&& \textbf{Decision-centric} & \textbf{100\%} & \textbf{1.10} & \textbf{4.20} \\
\midrule
\multirow{3}{*}{4}
& \multirow{3}{*}{---}
& Retry baseline       &   0\% & 6.00 & 6.00 \\
&& Prompt-based clarify &  10\% & 2.90 & 6.00 \\
&& \textbf{Decision-centric} & \textbf{100\%} & \textbf{0.10} & \textbf{2.50} \\
\bottomrule
\end{tabular}
\caption{Full results by missing-field count $k$ and ambiguity type.
  Granite~4 (micro), $T{=}6$, $N{=}10$ per cell.
  ``Absent'': missing fields simply omitted.
  ``Unresolvable'': missing fields referenced but not inferable.}
\label{tab:appendix_full}
\end{table}

DC maintains 100\% on all absent scenarios and holds 100\% on all unresolvable scenarios except $k{=}3$ (where it still outperforms Prompt-Clarify's 60\% with far fewer turns).
The unresolvable condition consistently widens the gap: Prompt-Clarify drops to 50\%, 60\%, and 10\% at $k{=}2,3,4$ unresolvable, whereas DC never fails, because it continues clarifying until fields are explicitly confirmed rather than treating vague references as resolved.
The Retry baseline is unaffected by ambiguity type, confirming its failures are structural (never clarifying) rather than ambiguity-driven.

\subsection{Cross-Model Analysis (LLaMA~3)}
\label{app:cross_model}

The decision function is identical across models---the same 3-branch rule, the same estimator---so performance differences are attributable entirely to model-specific component behavior, not control logic.
Table~\ref{tab:cross_model_full} gives the full per-scenario breakdown.
\begin{itemize}[leftmargin=*,itemsep=2pt]
  \item \textbf{DC~(original) outperforms both baselines at every $k$.}
    Even before any fix, DC~(original) beats Retry and Prompt-Clarify at
    every missing-field count, confirming that the explicit decision layer
    transfers across model families.
  \item \textbf{The error pattern is non-monotonic.}
    Unlike Granite, where all approaches degrade monotonically with $k$,
    DC~(original) on LLaMA~3 reaches 100\% at $k{=}2$ and $k{=}4$ but only
    50\%--75\% at $k{=}1$, and Prompt-Clarify peaks at $k{=}2$ absent (90\%)
    before falling to 40\% at $k{=}3$ unresolvable.
    Because the decision layer is explicit, this irregularity can be
    localized to a specific component rather than treated as an opaque model effect.
  \item \textbf{DC~(constrained) recovers the gap.}
    A targeted fix to the question-generator prompt---leaving the estimator,
    policy, and executor unchanged---restores 100\% at all $k$ except
    $k{=}3$ unresolvable (90\%).
\end{itemize}

\begin{table}[h]
\centering
\small
\begin{tabular}{c l rrr rr rr}
\toprule
& & \multicolumn{2}{c}{Retry} & \multicolumn{2}{c}{Prompt-clarify} & DC (orig.) & DC (constrained) \\
\cmidrule(lr){3-4}\cmidrule(lr){5-6}
$k$ & Ambiguity & Success & Turns & Success & Turns & Success & Success \\
\midrule
0 & --- & 100\% & 1.00 &  70\% & 5.50 & 100\% & 100\% \\
\midrule
\multirow{2}{*}{1}
& Absent       &  10\% & 5.90 &  60\% & 5.50 &  50\% & 100\% \\
& Unresolvable &   0\% & 6.00 &  40\% & 5.90 & 100\% & 100\% \\
\midrule
\multirow{2}{*}{2}
& Absent       &   0\% & 6.00 &  90\% & 5.60 & 100\% & 100\% \\
& Unresolvable &   0\% & 6.00 &  60\% & 5.80 & 100\% & 100\% \\
\midrule
\multirow{2}{*}{3}
& Absent       &   0\% & 6.00 &  50\% & 5.90 & 100\% & 100\% \\
& Unresolvable &   0\% & 6.00 &  40\% & 6.00 &  80\% &  90\% \\
\midrule
4 & --- &   0\% & 6.00 &  30\% & 5.90 & 100\% & 100\% \\
\bottomrule
\end{tabular}
\caption{LLaMA~3~(8B) full per-scenario breakdown, $N{=}10$, $T{=}6$.
  DC~(orig.): default question-generator prompt.
  DC~(constrained): stricter prompt constraining question generation to missing fields only.
  The single change at $k{=}1$ absent (50\%~$\to$~100\%) isolates the broken component:
  only the question generator was changed; all other components are identical.
  Prompt-clarify uses 5.5--6.0 turns on every scenario, exhausting the budget
  in clarification rather than executing.}
\label{tab:cross_model_full}
\end{table}

\begin{figure}[h]
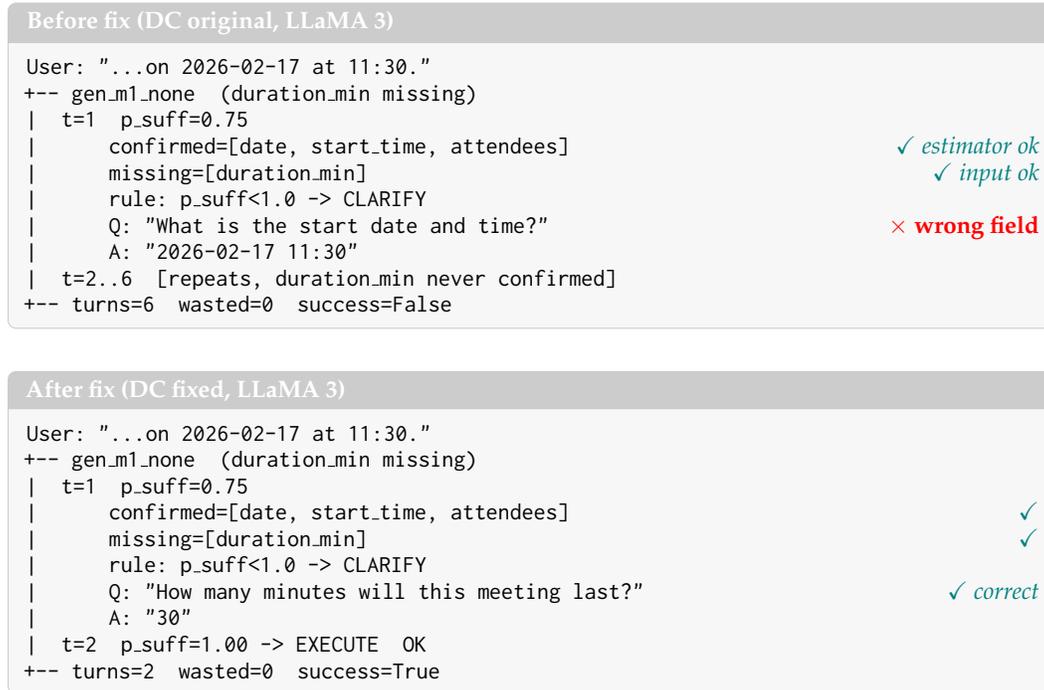

\centering
\small
\begin{tcolorbox}[colback=gray!6,colframe=gray!40,boxrule=0.4pt,
  title={\small\textbf{Before fix (DC original, LLaMA~3)}},fonttitle=\small,
  left=4pt,right=4pt,top=3pt,bottom=3pt]
\small\ttfamily
User: "...on 2026-02-17 at 11:30."\\
+-- gen\_m1\_none\ \ (duration\_min missing)\\
|\ \ t=1\ \ p\_suff=0.75\\
|\ \ \ \ \ \ confirmed=[date, start\_time, attendees]\hfill{\normalfont\itshape\color{teal}\checkmark\ estimator ok}\\
|\ \ \ \ \ \ missing=[duration\_min]\hfill{\normalfont\itshape\color{teal}\checkmark\ input ok}\\
|\ \ \ \ \ \ rule: p\_suff<1.0\ ->\ CLARIFY\\
|\ \ \ \ \ \ Q: "What is the start date and time?"\hfill{\normalfont\bfseries\color{red}$\times$\ wrong field}\\
|\ \ \ \ \ \ A: "2026-02-17 11:30"\\
|\ \ t=2..6\ \ [repeats, duration\_min never confirmed]\\
+-- turns=6\ \ wasted=0\ \ success=False
\end{tcolorbox}
\vspace{6pt}
\begin{tcolorbox}[colback=gray!6,colframe=gray!40,boxrule=0.4pt,
  title={\small\textbf{After fix (DC fixed, LLaMA~3)}},fonttitle=\small,
  left=4pt,right=4pt,top=3pt,bottom=3pt]
\small\ttfamily
User: "...on 2026-02-17 at 11:30."\\
+-- gen\_m1\_none\ \ (duration\_min missing)\\
|\ \ t=1\ \ p\_suff=0.75\\
|\ \ \ \ \ \ confirmed=[date, start\_time, attendees]\hfill{\normalfont\itshape\color{teal}\checkmark}\\
|\ \ \ \ \ \ missing=[duration\_min]\hfill{\normalfont\itshape\color{teal}\checkmark}\\
|\ \ \ \ \ \ rule: p\_suff<1.0\ ->\ CLARIFY\\
|\ \ \ \ \ \ Q: "How many minutes will this meeting last?"\hfill{\normalfont\itshape\color{teal}\checkmark\ correct}\\
|\ \ \ \ \ \ A: "30"\\
|\ \ t=2\ \ p\_suff=1.00\ ->\ EXECUTE\ \ OK\\
+-- turns=2\ \ wasted=0\ \ success=True
\end{tcolorbox}
\caption{Diagnostic trace for LLaMA~3 at $k{=}1$ (before and after fix).
  Three logged values pinpoint the broken component: $\hat{p}_\text{suff}=0.75$ is
  correct; \texttt{missing=[duration\_min]} is correct; yet the question asks about
  date and time---only the question generator is wrong.
  The fix targets only the question-generation prompt; the estimator, policy, and
  execution function are unchanged.}
\label{fig:llama_trace}
\end{figure}

\paragraph{Diagnosing the $k{=}1$ failure.}
DC~(original) achieves only 50\% at $k{=}1$ absent due to an instruction-following
failure in the question generator: given \texttt{missing\_fields=["duration\_min"]},
LLaMA~3 generates questions about date and time instead.
The estimator correctly returns $\hat{p}_\text{suff}=0.75$ with \texttt{duration\_min}
unconfirmed---the signal and policy are correct; only the question generator is broken.
Figure~\ref{fig:llama_trace} shows the before/after trace where three logged values
immediately pinpoint the component at fault.

The fix tightens the question-generation prompt without changing anything else.
LLaMA~3 has weaker instruction-following on constrained generation~\citep{zhou2023ifeval}
and drifts to asking about unrelated fields unless explicitly constrained:

\medskip
\noindent\textbf{Default prompt (Granite):}
\begin{quote}\small\ttfamily
Generate ONE natural question that asks the user for all the specified missing fields.
\end{quote}

\noindent\textbf{Constrained prompt (LLaMA~3):}
\begin{quote}\small\ttfamily
You must ask the user ONLY about the fields listed under "missing\_fields".\\
Do not ask about any other fields.\\
Generate ONE question covering exactly those missing fields.
\end{quote}

\noindent The estimator, policy, and execution function are identical in both cases.
Only this prompt string is model-dependent.

\paragraph{Explicit separation enables targeted repair.}
The diagnosis points directly to a minimal fix: strengthen the
question-generation prompt so that LLaMA~3 asks only about the listed
missing fields. No other component changes: the estimator, decision
policy, and execution function remain fixed. After this repair, DC
recovers to 100\% at $k{=}1$ and 95\% at $k{=}3$
(Table~\ref{tab:cross_model_main}).

This is the intended benefit of the modular architecture: failures are
attributable, repairs are surgical, and improvement is directly
measurable. Such component-level diagnosis is possible because the
decision layer is explicit; in prompt-based systems, there is no
comparable separation.

\paragraph{Remaining failure at $k{=}3$ unresolvable.}
The query ``Schedule a meeting with the usual team'' contains an unresolvable
attendee reference. The extractor marks \texttt{attendees=true} from the vague
reference---the first logged value points to the bug.
The executor copies the vague string literally, producing \texttt{["the usual team"]},
and validation fails.
The no-blind-retry constraint fires correctly, but the fallback confirmation question
does not recover because the user's ``Yes'' does not supply a specific name.
This is a distinct failure from the $k{=}1$ bug: the extractor and executor jointly
mishandle unresolvable references. Traceability again makes the cause immediate:
\texttt{confirmed\_fields} shows \texttt{attendees=true} at $t{=}1$ despite no
named person in the query, pointing directly to the extractor as the component to fix.


\section{Graph Disambiguation}
\label{app:graph_task}

\subsection{Experiment Setup}\label{app:graph_setup}

The knowledge graph contains 200 nodes (people), each described by five
categorical attributes:
\texttt{department} $\in$ \{Engineering, Marketing, Sales, Finance\},
\texttt{role} $\in$ \{Manager, Analyst, Engineer, Lead\},
\texttt{location} $\in$ \{New York, London, Tokyo, Berlin\},
\texttt{project} $\in$ \{Alpha, Beta, Gamma, Delta\}, and
\texttt{level} $\in$ \{L1, L2, L3, L4\}.
Edges connect pairs of nodes that share at least two attribute values.
To introduce realistic distractors, we add noisy edges equal to 15\% of the
base edge count between pairs sharing at most one attribute; these edges are
not distinguishable from regular edges before traversal.

\paragraph{Controlled subgraph for S5.}
S5 is a stress test designed to isolate a specific failure mode: correlated
belief update, where a failed traversal passively improves sufficiency through
elimination without any clarification turn.
This property requires a precise structural condition, i.e., a clarification that
leaves $\ge 5$ candidates including a minority group whose elimination is
triggered by a subsequent traversal failure.
At 200 nodes over a $4^5 = 1024$-combination attribute space, this pattern
does not arise naturally with sufficient regularity for controlled evaluation.
We therefore inject 10 nodes sharing
(\texttt{department=Engineering}, \texttt{project=Alpha}, \texttt{role=Manager})
with the profiles in Table~\ref{tab:s5_subgraph}, chosen to instantiate the
target structure exactly.
These nodes participate in the graph's edge construction identically to all
other nodes and are indistinguishable from random nodes by any approach.
S1--S4 are unaffected.

\begin{table}[h]
\centering
\small
\begin{tabular}{l l l l}
\toprule
Node & Location & Level & Role in S5 \\
\midrule
Target  & Tokyo  & L3 & Correct answer \\
D       & Tokyo  & L1 & Forced first visit; triggers elimination \\
E       & Tokyo  & L1 & Eliminated when D's \texttt{level=L1} is observed \\
F       & Tokyo  & L2 & Survives elimination \\
F2      & Tokyo  & L4 & Survives elimination \\
A       & London & L2 & Removed by \texttt{location=Tokyo} clarification \\
B       & London & L1 & Removed by \texttt{location=Tokyo} clarification \\
C       & London & L4 & Removed by \texttt{location=Tokyo} clarification \\
G       & Berlin & L2 & Removed by \texttt{location=Tokyo} clarification \\
H       & Berlin & L1 & Removed by \texttt{location=Tokyo} clarification \\
\bottomrule
\end{tabular}
\caption{Injected S5 subgraph. All 10 nodes share
  \texttt{department=Engineering}, \texttt{project=Alpha}, \texttt{role=Manager}.
  After clarifying \texttt{location=Tokyo}: 5 candidates remain (Target, D, E, F, F2).
  After visiting D: \texttt{level=L1} is minority (1 of 4 non-target peers $=$ 25\%),
  so E is eliminated, leaving 3 candidates.}
\label{tab:s5_subgraph}
\end{table}

\paragraph{Scenario Specifications}

We organize the scenarios into two tiers.
\textbf{S1--S3 are controls}: S1 validates the end-to-end pipeline; S2 isolates
$\hat{p}_\text{suff}$ in the absence of a correctness signal; and S3 isolates
$\hat{p}_\text{corr}$ in a setting with only one remaining alternative, where
all approaches can ultimately recover by exhaustion.
\textbf{S4--S5 are the main tests}: S4 probes the joint-belief setting, where
the optimal post-failure action depends on both signals simultaneously; S5
probes correlated belief updates, where a failed traversal improves sufficiency
through candidate elimination without requiring another clarification turn.

\begin{itemize}[leftmargin=*,itemsep=4pt]
  \item \textbf{S1 -- Clean (baseline).}
    4 of 5 attributes known; 1 candidate ($\hat{p}_\text{suff}{=}1.0$); $T{=}5$.
    Optimal: execute $\to$ accept. Sanity check.

  \item \textbf{S2 -- Ambiguous ($\hat{p}_\text{suff}$ only).}
    2 of 5 attributes known; 13 candidates ($\hat{p}_\text{suff}{=}0.077$); $T{=}5$.
    Optimal: clarify $\times$2 $\to$ execute $\to$ accept.
    Isolates $\hat{p}_\text{suff}$: the decision problem is purely clarify-or-execute.

  \item \textbf{S3 -- Unreliable edge ($\hat{p}_\text{corr}$ only).}
    4 of 5 attributes known; 2 candidates; forced first visit to a decoy
    with a \emph{unique} hidden profile (no elimination); $T{=}5$.
    Optimal: execute $\to$ detect low $\hat{p}_\text{corr}$ $\to$ backtrack $\to$ accept.
    Isolates $\hat{p}_\text{corr}$: unique profile ensures no elimination fires.

  \item \textbf{S4 -- Orthogonal joint belief.}
    2 of 5 attributes known; 10 candidates; forced decoy with unique hidden
    profile; $T{=}6$.
    After the decoy fails (low $\hat{p}_\text{corr}$), the action depends on $\hat{p}_\text{suff}$:
    with many untried candidates, backtracking is wasteful---clarify instead.
    The same low $\hat{p}_\text{corr}$ produces backtrack in S3 (high $\hat{p}_\text{suff}$,
    few alternatives) and clarify in S4 (low $\hat{p}_\text{suff}$, many alternatives).

  \item \textbf{S5 -- Correlated belief update.}
    3 of 5 attributes known; 12 initial candidates; forced first execution to decoy D; $T{=}5$.
    The defining feature: $\hat{p}_\text{suff}$ rises at $t{=}2$ from a failed traversal
    that triggers elimination, not from a clarification.
\end{itemize}

\subsection{Decision-Centric Approach Implementation}\label{app:graph_dc}

\paragraph{Decision Policy.}

DC uses a fixed policy over the explicit signals
$\hat p_{\mathrm{suff}}$, $\hat p_{\mathrm{corr}}$, and the current search state.
The policy is specified a priori rather than fit from data.
We use $\tau_{\mathrm{suff}} = 0.4$, so execution begins once the remaining
candidate set is sufficiently small (e.g., $1/2 = 0.5 > \tau_{\mathrm{suff}}$,
whereas $1/3 = 0.33 < \tau_{\mathrm{suff}}$), and
$\theta_{\mathrm{corr}} = 0.5$, the midpoint of $[0,1]$, as the accept/reject
cutoff after traversal.

Operationally, the policy distinguishes between two states.
Before traversal, it decides whether the current description is specific enough
to act on; after traversal, it decides whether to accept the visited node,
backtrack, or request further clarification.

\begin{tcolorbox}[colback=gray!6,colframe=gray!40,boxrule=0.4pt,
  title={\small\textbf{Graph Disambiguation Policy}},fonttitle=\small,
  left=4pt,right=4pt,top=3pt,bottom=3pt]
\small\ttfamily
\textbf{Input:}\ $\hat{p}_\text{suff}$,\ $\hat{p}_\text{corr}$,\ just\_traversed,\ n\_untried,\ n\_hidden,\ turns\\[2pt]
\begin{tabular}{@{}l@{\enspace}l@{\enspace}l@{}}
if just\_traversed: & & \\
\ \ \ \ if p\_corr >= theta\_corr: & -> & ACCEPT \\
\ \ \ \ if n\_untried == 0: & -> & ACCEPT \quad \# no alternatives \\
\ \ \ \ if p\_suff < tau\_suff and turns > 2: & -> & CLARIFY \quad \# too many candidates \\
\ \ \ \ elif n\_untried > 0: & -> & BACKTRACK \\
\ \ \ \ else: & -> & CLARIFY \quad \# exhausted \\
else: \quad \# pre-traversal & & \\
\ \ \ \ if p\_suff >= tau\_suff: & -> & EXECUTE \\
\ \ \ \ if n\_hidden <= 1: & -> & EXECUTE \quad \# last-hidden-attr rule \\
\ \ \ \ else: & -> & CLARIFY \\
\end{tabular}
\end{tcolorbox}

The final pre-traversal rule handles cases in which only one attribute remains
unobserved. In that regime, another clarification would further narrow the
candidate set, but executing can also be efficient: a failed traversal reveals
a concrete mismatch and supports elimination through backtracking. In S5, this
rule favors execution once only \texttt{level} remains hidden, allowing the
system to reduce the candidate set through the traversal outcome rather than
spending an additional clarification turn.

\paragraph{Correctness Estimator}

$\hat{p}_\text{corr}$ is estimated from the visited node's hidden-attribute
profile relative to the active peer set:
\begin{enumerate}[noitemsep]
  \item For each hidden attribute $k$, compute the fraction of active peer
    candidates (excluding the visited node and previously rejected nodes) that
    share the visited node's value.
  \item Average these fractions across hidden attributes.
  \item Penalize similarity to prior rejections: if the visited node matches a
    previously rejected hidden profile on at least $50\%$ of hidden attributes,
    subtract $0.15$--$0.35$.
  \item Clamp the result to $[0.05, 0.95]$.
\end{enumerate}

\subsection{Prompt Baseline Implementation}
\label{app:graph_prompts}

Both prompt approaches receive the same per-turn JSON payload:
\texttt{query}, \texttt{known\_attributes}, \texttt{unknown\_attributes},
\texttt{n\_candidates\_matching}, \texttt{n\_untried\_candidates},
\texttt{last\_visited\_profile}, \texttt{n\_rejected}, \texttt{turn}, \texttt{budget}.
They differ only in the system prompt.

\paragraph{Prompt-clarify system prompt (temperature~$0.5$).}

\begin{quote}\small\ttfamily
You are searching for a specific person in an organization of 200 people.
Each person has 5 attributes: department, role, location, project, level.\\[4pt]
You have a partial description and access to search results showing how many
people match it.\\[4pt]
At each turn, choose ONE action:\\
\quad \{"action": "clarify", "attribute": "<attr>"\}\\
\quad\quad --- ask the user for a missing attribute to narrow down candidates\\
\quad \{"action": "execute"\}\\
\quad\quad --- visit a candidate node to check if it's the target\\
\quad \{"action": "backtrack"\}\\
\quad\quad --- reject the last visited node and try a different candidate\\
\quad\quad\quad (only if you've visited a node that didn't look right AND alternatives exist)\\[4pt]
Budget is limited --- don't waste turns.\\
Available attributes to ask about: only those NOT already in the description.
\end{quote}

\paragraph{Prompt (w/ policy) system prompt.}
Identical to the above, with the following block added before the final line:

\begin{quote}\small\ttfamily
\textbf{[policy block, absent from prompt-clarify]}\\[2pt]
Consider:\\
\quad With many candidates (n\_candidates\_matching is large), clarifying\\
\quad first is usually safer.\\
\quad After visiting a wrong node, you can backtrack (try another) or\\
\quad clarify (narrow the field first, then try again).
\end{quote}

This block specifies the decision logic that DC makes explicit.
It improves performance when the correct action can be read directly from the
observable candidate state: +9pp on S2, where the system should clarify before
executing into a pool of 13 candidates, and +5pp on S4, where the correct
post-decoy action is to clarify rather than backtrack.
On S5, both variants remain at 35\%. The bottleneck there is not failure to
reason about the candidate count, but failure to observe that traversal has
already reduced the pool through elimination. Because that updated state is not
exposed in the decision-time payload, the LLM cannot act on it.

\subsection{Full Results by Scenario}

\begin{table}[h]
\centering
\small
\begin{tabular}{@{}l l rrrr@{}}
\toprule
Scenario & Approach & Success & Wasted & Clarify & Backtrack \\
\midrule
\multirow{4}{*}{\shortstack[l]{S1: Clean}}
& Retry              & 100\% & 0.00 & 0.00 & 0.00 \\
& Prompt             & 100\% & 0.00 & 0.00 & 0.00 \\
& Prompt (w/ policy) & 100\% & 0.00 & 0.00 & 0.00 \\
& DC                 & 100\% & 0.00 & 0.00 & 0.00 \\
\midrule
\multirow{4}{*}{\shortstack[l]{S2: Ambiguous\\($\hat{p}_\text{suff}$ only)}}
& Retry              &  45\% & 3.70 & 0.00 & 0.00 \\
& Prompt             &  85\% & 1.35 & 1.55 & 0.00 \\
& Prompt (w/ policy) &  95\% & 0.60 & 2.00 & 0.00 \\
& DC                 & 100\% & 1.00 & 2.00 & 1.00 \\
\midrule
\multirow{4}{*}{\shortstack[l]{S3: Unreliable\\($\hat{p}_\text{corr}$ only)}}
& Retry              & 100\% & 1.00 & 0.00 & 0.00 \\
& Prompt             & 100\% & 1.00 & 0.00 & 0.00 \\
& Prompt (w/ policy) & 100\% & 1.00 & 0.00 & 0.00 \\
& DC                 & 100\% & 1.00 & 0.00 & 1.00 \\
\midrule
\multirow{4}{*}{\shortstack[l]{S4: Orthogonal\\(joint, $T{=}6$)}}
& Retry              &  65\% & 4.20 & 0.00 & 0.00 \\
& Prompt             &  95\% & 1.05 & 1.50 & 0.00 \\
& Prompt (w/ policy) & 100\% & 0.40 & 1.95 & 0.00 \\
& DC                 & 100\% & 1.00 & 2.00 & 1.00 \\
\midrule
\multirow{4}{*}{\shortstack[l]{S5: Correlated\\(coupled update)}}
& Retry              &  60\% & 3.50 & 0.00 & 0.05 \\
& Prompt             &  35\% & 4.10 & 0.05 & 0.10 \\
& Prompt (w/ policy) &  35\% & 4.05 & 0.20 & 0.10 \\
& DC                 & 100\% & 0.60 & 2.00 & 0.00 \\
\bottomrule
\end{tabular}
\caption{Full graph disambiguation results ($N{=}20$, structural $\hat{p}_\text{corr}$ estimator).
  \emph{Wasted}: traversals to wrong candidates per run.
  \emph{Clarify}: clarification turns per run.
  \emph{Backtrack}: backtrack actions per run.}
\label{tab:graph_full}
\end{table}

\subsection{DC Belief Traces}
\label{app:graph_trace}

\paragraph{S4 trace (orthogonal joint belief).}

\begin{verbatim}
Query: "Find the person whose role is Manager, at level L2."
  [10 candidates, 2 known attrs, forced decoy]
  t=1 p_suff=0.100 p_corr=1.000 cands=10  -> CLARIFY
      asked=department, answer=Marketing
  t=2 p_suff=0.333 p_corr=1.000 cands=3   -> CLARIFY
      asked=location, answer=Tokyo
  t=3 p_suff=0.500 p_corr=1.000 cands=2   -> EXECUTE
      -> decoy  correct=False
         p_corr=0.375 (outlier hidden profile)
         [no elimination: unique hidden profile]
  t=4 p_suff=0.500 p_corr=0.375 untried=1 -> BACKTRACK
      -> target  correct=True  p_corr=0.900
  t=5 p_suff=0.500 p_corr=0.900 untried=0 -> ACCEPT
  turns=4  clarify=2  backtrack=1  success=True
\end{verbatim}

At t=4, the joint state is (high $\hat{p}_\text{suff}$, low $\hat{p}_\text{corr}$)
$\to$ backtrack. In S2 with 13 candidates, the same low $\hat{p}_\text{corr}$
would produce clarify, i.e., same signal, different action, driven by the joint state.

\paragraph{S5 trace (correlated belief update).}

\begin{verbatim}
Query: "Find the person in Engineering, project Alpha, role Manager."
  [12 candidates, 3 known attrs]
  t=1 p_suff=0.083 p_corr=1.000 cands=12  -> CLARIFY
      asked=location, answer=Tokyo
      [5 candidates remain: Target, D, E, F, F2]
  t=2 p_suff=0.200 p_corr=1.000 cands=5
      [1 hidden attr remaining: level]
      last-hidden-attr rule fires             -> EXECUTE
      -> S5_Decoy_D (level=L1)  correct=False
         p_corr=0.250  (L1 is 25% of peers)
         [elimination: S5_Node_E (level=L1) eliminated]
         [p_suff: 0.200 -> 0.333  (5 -> 3 candidates)]
  t=3 p_suff=0.333 p_corr=0.250 untried=2  -> BACKTRACK
      -> S5_Node_F (level=L2)  correct=False
  t=4 p_suff=0.333 p_corr=0.250 untried=1  -> BACKTRACK
      -> S5_Target (level=L3)  correct=True
         p_corr=0.950
  turns=4  clarify=1  backtrack=2  elim=1  success=True
\end{verbatim}

The elimination at t=2 raises $\hat{p}_\text{suff}$ from 0.200 to 0.333 without
a clarification turn.
Prompt-clarify and prompt (w/ policy) both see the updated candidate count at t=3
but have already committed their t=2 action without knowing the elimination occurred.
In 65\% of runs they spend t=2 on clarification (asking about \texttt{level}, which
DC already learned structurally from the failed traversal) or backtrack randomly
without exploiting the narrowed pool, exhausting the budget before finding the target.

\section{Modular Signals for Retrieval Control with Diagnosable Feedback}
\label{app:rag}

\subsection{Experimental Setup}\label{rag:setup}

\paragraph{Corpus and passage construction.}
We load 2000 items from the Natural Questions validation set~\citep{kwiatkowski2019natural},
retaining only items with both a short-answer annotation and a long-answer paragraph span.
Each item's passage is the long-answer paragraph extracted from the NQ document token sequence
(HTML tags filtered), guaranteeing the gold answer string appears in the passage.
A BM25 index is built over these 2000 annotated passages.

\paragraph{Difficulty buckets.}
Buckets are defined by the BM25 rank of each item's \emph{annotated passage} under the original question---not by string matching over retrieved text.
\begin{itemize}[noitemsep]
  \item \textbf{Easy}: annotated passage ranks in BM25 top-$k$ at round~0 ($k{=}3$).
  \item \textbf{Medium}: annotated passage not in top-$k$ at round~0, but retrieved by round~2 (top-$2k$ or top-$3k$).
  \item \textbf{Hard}: annotated passage never retrieved within budget (outside top-$3k$).
\end{itemize}
This definition is independent of all signals: DC-Dense and DC-LLM never see the passage index used for bucketing.
Retrieval states (accumulated passages at $k{=}3,6,9$) are precomputed once and shared across all approaches.

\subsection{Baselines Implementations} \label{rag:signals}

\paragraph{Prompt.}
The same LLM as DC-LLM reads the current passages and directly outputs
$\{\texttt{stop},\texttt{expand\_k}\}$ with a free-text reason string.
No numeric signal is logged.
The system prompt instructs the model not to use its own knowledge, identical restriction to DC-LLM, so the only structural difference is whether the sufficiency assessment is externalized.

\paragraph{DC-Dense.}
Max cosine similarity between the question embedding and passage embeddings
(\texttt{sentence-transformers/all-MiniLM-L6-v2}),
linearly normalised to $[0,1]$.
No LLM calls.

\paragraph{DC-LLM.}
LLM answerability judge with the prompt:
\emph{``Assess whether the passages EXPLICITLY contain the answer.
Do NOT use your own knowledge.
Output JSON: \{answerable: true or false, confidence: 0--1\}''},
temperature~0.1.
$\hat{p}_\text{llm} = \text{confidence}$ if answerable, else $1 - \text{confidence}$,
then linearly normalised to $[0,1]$.

\paragraph{DC-Composite.}
$\hat{p} = \alpha \hat{p}_\text{dense} + (1{-}\alpha)\hat{p}_\text{llm}$, with $\alpha{=}0.4$.

All DC variants apply the controller: stop if $\hat{p} \geq \tau$, where $\tau{=}0.8$, else expand (forced stop at budget).
A robustness sweep over $\tau \in \{0.5,\ldots,0.9\}$ and $\alpha \in \{0.2,\ldots,0.6\}$ is reported in Appendix~\ref{app:rag_sweep}.

\subsection{Threshold Robustness}
\label{app:rag_sweep}

The parameters $\alpha{=}0.4$ and $\tau{=}0.8$ were selected on a separate
validation set (30 questions, 10 per bucket) prior to running the test set (150 questions, 50 per bucket).
Table~\ref{tab:rag_sweep} reports the sweep on the held-out test set,
computed offline by replaying the saved per-round signal traces without re-running any model calls.
Easy-bucket success remains 100\% throughout; the chosen setting is conservative
rather than optimized---higher $\tau$ or $\alpha$ further improves medium-bucket
success but at the cost of more retrieval rounds on easy questions.
The test-set results are consistent with what was observed on the validation set,
confirming the setting was not cherry-picked.

\begin{table}[h]
\centering
\small
\begin{tabular}{cc rr r}
\toprule
$\alpha$ & $\tau$ & Easy Succ. & Easy RR & Medium Succ. \\
\midrule
0.4 & 0.5 & 100\% & 0.66 & 78\% \\
0.4 & 0.6 & 100\% & 0.66 & 78\% \\
0.4 & 0.7 & 100\% & 0.70 & 82\% \\
\textbf{0.4} & \textbf{0.8} & \textbf{100\%} & \textbf{0.84} & \textbf{88\%} \\
0.4 & 0.9 & 100\% & 1.16 & 92\% \\
\midrule
0.2 & 0.8 & 100\% & 0.66 & 78\% \\
0.3 & 0.8 & 100\% & 0.70 & 84\% \\
\textbf{0.4} & \textbf{0.8} & \textbf{100\%} & \textbf{0.84} & \textbf{88\%} \\
0.5 & 0.8 & 100\% & 0.92 & 90\% \\
0.6 & 0.8 & 100\% & 1.00 & 90\% \\
\bottomrule
\end{tabular}
\caption{Offline threshold sweep for DC-Composite on the held-out test set ($N{=}50$ per bucket).
  Parameters were selected on a separate validation set ($N{=}30$) before the test set was run.
  Top block: varying $\tau$ at $\alpha{=}0.4$.
  Bottom block: varying $\alpha$ at $\tau{=}0.8$.
  Bold row is the paper's setting.
  Easy success is 100\% throughout; the chosen operating point sits below the
  recoverable peak, trading a small gain in medium-bucket success for fewer
  wasted retrieval rounds on easy questions.}
\label{tab:rag_sweep}
\end{table}

\subsection{Signal Attribution and Feedback Loop}
\label{app:rag_attribution}

A benefit of the decision-centric design is that failures can be attributed
directly to the sufficiency signals. Because each DC run logs
$\hat{p}_\text{dense}$ and $\hat{p}_\text{llm}$ at every round, failed episodes
can be analyzed offline without re-running the model.

\paragraph{Attribution procedure.}
For each failed DC-Composite episode, we identify the round at which the wrong
action was taken and inspect the logged signal values at that round. We assign
the failure to one of four categories:
\begin{itemize}[noitemsep]
  \item \textbf{Early stop, dense-caused}: $\hat{p}_\text{dense}$ is high and
        dominant, so dense similarity overestimates sufficiency.
  \item \textbf{Early stop, LLM-caused}: $\hat{p}_\text{llm}$ is high and
        dominant, so the answerability judge is overconfident.
  \item \textbf{Early stop, both signals high}: both signals support
        \texttt{stop}, but both are wrong.
  \item \textbf{Corpus gap (over-expand)}: the controller expands to budget, but
        the gold passage is never retrieved, so signal improvement alone cannot
        fix the failure.
\end{itemize}

\paragraph{Attribution results.}
Table~\ref{tab:attribution} reports the breakdown for DC-Composite
($\tau{=}0.8$, $\alpha{=}0.4$). The medium-bucket failures are entirely cases
where both signals are confidently wrong. By contrast, hard-bucket failures are
almost entirely corpus gaps: the relevant passage never appears within budget,
so the controller has no signal-level remedy available.

\begin{table}[h]
\centering
\small
\begin{tabular}{l l rr}
\toprule
Bucket & Failure type & $N$ & \% \\
\midrule
\multirow{2}{*}{Medium} &
  Early stop, both signals high & 3 & 100\% \\
& \emph{(3 total failures)} & & \\
\midrule
\multirow{2}{*}{Hard} &
  Corpus gap (over-expand) & 40 & 98\% \\
& Early stop, both signals high & 1 & 2\% \\
& \emph{(41 total failures)} & & \\
\bottomrule
\end{tabular}
\caption{Failure attribution for DC-Composite on the test set. Medium failures
are dominated by cases in which both signals support an incorrect early
\texttt{stop}. Hard failures are dominated by corpus gaps, where the gold
passage is never retrieved within budget.}
\label{tab:attribution}
\end{table}

\paragraph{Implications for signal design.}
The attribution suggests different next steps for the two buckets.

\textbf{Medium: both signals high (3/3).}
Here both DC-Dense and DC-LLM support \texttt{stop} incorrectly: dense
similarity is high because the retrieved passages are topically related, and
the LLM judge is overconfident even though the answer is not explicitly stated.
Reweighting the existing signals is therefore unlikely to help. What is needed
is a signal of a different type, for example:
\begin{itemize}[noitemsep]
  \item \emph{Passage highlighting}: require the model to quote the exact answer
        span verbatim.
  \item \emph{Answer-type matching}: check whether the passage contains an
        entity of the expected type (person, date, location, number).
  \item \emph{Token-level overlap}: measure lexical overlap between expected
        answer tokens and the passage text.
\end{itemize}

\textbf{Hard: corpus gap (40/41).}
Here the main limitation is retrieval, not signal quality. Improving
$\hat{p}$ will not help if the relevant passage never enters the candidate set.
The appropriate remedy is either a stronger retriever or a corpus-exhaustion
signal that recognizes diminishing returns and stops early rather than using the
full expansion budget.

\paragraph{Feedback loop.}
This attribution procedure instantiates the feedback loop described in
Section~\ref{sec:rag}. A failed outcome identifies the round where the wrong
action was taken; the logged traces reveal which signals supported that action;
the attribution pattern suggests what kind of signal is missing; and candidate
signals can then be evaluated offline by replaying the saved traces. The
controller remains unchanged throughout.

\subsection{Prompt Failure Analysis}
\label{app:rag_prompt_failures}

To analyze the prompt baseline, we log the free-text reason string produced on
each turn. On the medium bucket, 28 of 44 failures (64\%) fall into a single
dominant pattern: the model explicitly states that the current passages do not
contain the answer, yet still chooses \texttt{stop}. The remaining failures are
split across two smaller categories. Table~\ref{tab:prompt_failures} shows
representative examples.

This pattern is important because it isolates the core weakness of implicit
control. The model often makes the correct assessment in natural language, but
that judgment is not bound to the action it outputs. In other words, the
failure is not necessarily one of recognition, but of mapping recognition to
control.

\begin{table}[h]
\centering
\small
\begin{tabular}{p{1.8cm} p{4.2cm} p{6.2cm}}
\toprule
Type & Question / Gold answer & Prompt reason $\to$ action \\
\midrule
\textbf{Type~1} \newline
\emph{Correct assessment, wrong action} \newline
(28/44) &
  ``where did Iowa State football coach come from'' / \emph{Massillon, Ohio} &
  \emph{``Passages do not explicitly contain the answer\ldots''} $\to$ \texttt{stop} \\[4pt]
&
  ``what does it mean to fuse the spine'' / \emph{neurosurgical technique\ldots} &
  \emph{``Passages do not explicitly contain the meaning\ldots''} $\to$ \texttt{stop} \\
\midrule
\textbf{Type~2} \newline
\emph{Latches onto wrong passage} \newline
(13/44) &
  ``who are the two teams in Super Bowl 2018'' / \emph{Philadelphia Eagles} &
  \emph{``Passage 1 lists the teams as the Steelers and Patriots.''} $\to$ \texttt{stop} \\[4pt]
&
  ``where was Walking Tall filmed'' / \emph{Squamish, BC} &
  \emph{``Passage 1 mentions filming in Astoria, Oregon.''} $\to$ \texttt{stop} \\
\midrule
\textbf{Type~3} \newline
\emph{Other} \newline
(3/44) &
  ``how many Super Bowl games has the Patriots played'' / \emph{ten} &
  \emph{``Passages mention the Patriots have played in 9 Super Bowls.''} $\to$ \texttt{stop} \\
\bottomrule
\end{tabular}
\caption{Prompt failure modes on medium questions (test set, $N{=}44$ failures).
  Type~1 is the dominant pattern (64\%): the model's own reason string acknowledges
  the passages are insufficient, yet the fused call still outputs \texttt{stop}.
  This illustrates the core limitation of implicit control: a correct internal
  assessment cannot be acted upon when it is not externalized.}
\label{tab:prompt_failures}
\end{table}

\end{document}